\documentclass[conference]{IEEEtran}
\IEEEoverridecommandlockouts
\usepackage{cite}
\usepackage{amsmath,amssymb,amsfonts}
\usepackage{algorithmic}
\usepackage{graphicx}
\usepackage{textcomp}
\usepackage{xcolor}
\usepackage{hyperref}
\usepackage{booktabs}
\def\BibTeX{{\rm B\kern-.05em{\sc i\kern-.025em b}\kern-.08em
    T\kern-.1667em\lower.7ex\hbox{E}\kern-.125emX}}
\begin{document}

\title{Diffeomorphic Motion Priors For Robust Cell Tracking in Dense, Non-Linear Migration Regimes: A Kalman-LDDMM framework}

\author{
\IEEEauthorblockN{Bishal Neupane\textsuperscript{1} (Corresponding Author)}
\IEEEauthorblockA{Email: neupanebishal039@gmail.com}
\and
\IEEEauthorblockN{Niharika Sharma\textsuperscript{2}}
\IEEEauthorblockA{Email: sh.niharika08@gmail.com}}

\maketitle

\begin{abstract}
The study of automated cell tracking in time-lapse microscopy sequences enables the systematic analysis of red blood cell (RBC) hemodynamics, rheology, and microvascular transit. Tracking pipelines in high-throughput microfluidic systems are largely constrained by two problems: reliable state estimation under severe shape deformation and robust data association in dense flow fields. Existing methods based on nearest-neighbor and constant-velocity motion models fail in densely packed populations, where hydrodynamic interactions and shear forces induce non-linear, non-rigid trajectories. In this work, we present a principled tracking framework that couples a Kalman filter for recursive Bayesian state estimation with a diffeomorphic deformation field estimated via Large Deformation Diffeomorphic Metric Mapping (LDDMM) to construct an image-informed flow prior. The LDDMM registration solves a geodesic flow optimization over the group of diffeomorphisms to estimate a smooth, invertible, topology-preserving deformation between consecutive frames as they squeeze through microchannels. Unlike nucleated somatic cells, mature RBCs are non-proliferative and move passively; thus, LDDMM is leveraged exclusively to map non-rigid morphology shifts across frames rather than topological division events. By integrating this fluid-deformation motion prior to bias state prediction, our approach substantially improves Multiple Object Tracking Accuracy (MOTA) for densely populated sequence(Channel with Slits dataset) from 33.8\% to 55.3\% and reduce identity switched by over 44\% relative to a nearest-neighbor baseline, while maintaining identical localization precision (MOTP), and for less densely populated sequence(U-Shaped Channel dataset), MOTA score improved from 56.6\% to 58.0\% and reduced identity switches by 33.4\%, without any MOTP changes. As the trajectory of dynamics shift towards severe, non-linear and dense migration, the performance gains by the diffeomorphic prior becomes significantly more pronounced.\cite{b31}.
\end{abstract}


\section*{Statements and Declarations}
All authors certify that they have no affiliations with or involvement in any organization or entity with any financial interest or non-financial interest in the subject matter or materials discussed in this manuscript.

\section{Introduction}
Tracking cell trajectories in time-lapse microscopy enables measurement of bio-physical quantities including migration speed, directionality persistence, division symmetry, and cell-cycle duration, which are invisible in static imagery and inaccessible through population-level assays \cite{b23,b24}. The application of cell-tracking mechanism span a broad spectrum: from understanding wound healing and tumor invasion on a cellular scale to characterizing embryonic morphogenesis and immune cell surveillance. In multiple scenarios, tracking accuracy directly impacts the quality of biological conclusions.

Despite decades of research, automated cell tracking in densely packed populations has remained an open challenge. Cells may undergo rapid non-linear motion, undergo division, temporarily occlude one another, and leave or enter the field of view. Segmentation-based pipelines, which first detect cell bodies and then link detections across frames, must solve a combinatorial assignment problem at each time step while contending with detector errors. The difficulty is compounded in crowded sequences where the inter-cell spacing is comparable to the cell diameter, making association ambiguous under simple distance-based criteria.

\subsection{Limitations of Existing Approaches}
Existing classical approaches, including the linear assignment problem (LAP) framework of Jaqaman et al. \cite{b23}, achieve reliable tracking in sparse populations but assume pairwise distances as the sole association cost, without exploiting the rich motion information encoded in the image intensities themselves. Nearest-neighbor tracking algorithms show a similar limitation: under the constant-velocity assumption, the predicted position for cell $k$ is $\hat{\mathbf{x}}_{t+1} = \hat{\mathbf{x}}_t + \dot{\hat{\mathbf{x}}}_t \Delta t$. This assumption is not enough, when cell motion is  governed by collective tissue flows, substrate gradients, or active force generation that deviates from linear kinematics.

The Simple Online and Real-time Tracking (SORT) framework \cite{b22} augments nearest-neighbor association with a Kalman filter and Hungarian algorithm, which has shown a strong performance achievement on rigid-object tracking benchmarks. However, it still relies on a constant-velocity prior and does not exploit image content to refine motion predictions. In the context of cell tracking, where the image itself encodes the flow of cellular material, this represents a fundamental omission.

More recent approaches employ deep learning -- graph neural networks, recurrent architectures, and transformer-based association modules -- to learn motion models and appearance embeddings from data \cite{b21}. While these achieve state-of-the-art results on the Cell Tracking Challenge benchmarks \cite{b21}, they require large annotated training sets, which are expensive to produce in biological microscopy contexts and may not generalize across imaging modalities or cell types.

The CellTracker baseline evaluated in this work demonstrates the limitations of naive multi-object tracking in a dense 150-frame sequence, achieving only 33.8\% MOTA with 2793 identity switches -- reflecting the severe ambiguity inherent in per-frame nearest-neighbor association without principled motion models.

\subsection{Proposed Approach}
This work proposes a tracking framework that addresses the motion-modeling gap by coupling two well-founded mathematical theories: Kalman filtering for recursive Bayesian state estimation, and Large Deformation Diffeomorphic Metric Mapping (LDDMM) for image-driven dense motion estimation. The core insight is that the displacement between two consecutive frames is encoded in image intensities $I_t$ and $I_{t+1}$ as a spatially coherent deformation. By registering adjacent frames through a diffeomorphism -- a smooth, invertible, topology-preserving map -- the pixel-level displacement of each cell centroid can be estimated without requiring learned representations. This LDDMM-estimated displacement is then used to bias the Kalman prediction distribution toward the image-derived motion, the LDDMM displacement enters the prediction as an exogenous drift term / pseudo-measurement of a naive constant-velocity model with a structured, spatially-varying informative prior.

We further investigate two image fidelity criteria within the LDDMM optimization:
\begin{itemize}
    \item \textbf{MSE}: The classical $L^2$ pixel-difference criterion, providing a quadratic energy surface with closed-form gradients.
    \item \textbf{Sinkhorn}: An entropic-regularized optimal transport cost computed via the Sinkhorn algorithm \cite{b11,b12}, providing a geometrically richer similarity measure based on the Wasserstein geometry of image intensity distributions, with theoretical guarantees of bias correction via the Sinkhorn divergence formulation \cite{b13}.
\end{itemize}
The two variants allow us to empirically assess whether the theoretical advantages of optimal-transport-based image similarity materialize in the downstream tracking quality on this dataset.

\section{Dataset}
To evaluate the performance of our experiment, we utilize the FourSlits subset of the curated red blood cell tracking dataset introduced by Cimr\'ak et al. \cite{b31}. This benchmark dataset consists of experimental video sequence detailing the hydrodynamic flow of red blood cells (RBCs) through microfluidic channels. The dataset comprises 150 FourSlits annotated frames containing approximately 6{,}000 ground-truth bounding-box detections. It also incorporates simulated 2D fluid velocity fields, offering a realistic environment to assess target trajectory persistence and multi-object tracking accuracy in dynamic biological flows.

\section{Methodology}

\subsection{Problem Formulation}

Let $\mathcal{I} = \{I_t\}_{t=1}^{T}$ denote a sequence of $T$ microscopy image frames, where each frame $I_t : \Omega \subset \mathbb{R}^d \rightarrow \mathbb{R}$ maps a spatial domain $\Omega$ to pixel intensities, with $d \in \{2,3\}$. At each time step $t$, we observe a set of $N_t$ detected cell centroids:

$$\mathcal{D}_t = \left\{ \mathbf{x}_t^{(i)} \in \mathbb{R}^d \mid i = 1, \ldots, N_t \right\}$$

The cell tracking objective is to associate detections across frames, producing a set of trajectories $\mathcal{T} = \{\tau_k\}_{k=1}^{K}$, where each trajectory $\tau_k = \{(t, \mathbf{x}_t^{(k)})\}$ encodes the spatio-temporal path of cell $k$. The core challenge is twofold: (1) \textbf{state estimation} -- predicting where a cell will be at time $t+1$ given its history -- and (2) \textbf{data association} -- matching predicted states to observed detections under potential occlusion, division, and appearance/disappearance.

The proposed framework addresses these challenges by coupling a Kalman filter for linear state estimation with a deformation field computed via Large Deformation Diffeomorphic Metric Mapping (LDDMM), forming a principled probabilistic motion model.

\subsection{Cell Detection and Segmentation via Cellpose}
Prior to tracking, each frame $I_t$ must be processed to localize individual cells and extract their centroids, forming the detection set $\mathcal{D}_t$. We employ Cellpose\cite{b31}, a generalist instance segmentation model specifically designed for biological cell imagery as detector. Cellpose is built upon a U-Net-style encoder–decoder convolutional neural network that predicts, for each pixel, a horizontal and vertical gradient of a simulated diffusion process whose fixed point is the cell centroid — a representation referred to as the **spatial flow field**. The centroid of cell $i$ at frame $t$ is then computed as the intensity-weighted centre of mass of its mask:

$$\mathbf{x}_t^{(i)} = \frac{\sum_{\mathbf{p} \in \mathcal{M}_t^{(i)}} I_t(\mathbf{p})\,\mathbf{p}}{\sum_{\mathbf{p} \in \mathcal{M}_t^{(i)}} I_t(\mathbf{p})}$$

Cell pose is pre-trained on a curated, diverse collection of microscopy images spanning multiple cell types and modalities. This makes it applicable to varied microscopy dataset, where no previous target-specific models have been developed. It's generalist architecture eliminates the need for task-specific segmentation model development and helps to convoy morphological information such as cell area and shape. In this work, only the centroid $\mathbf{x}_t^{(i)}$ derived from each instance mask.

\subsection{State-Space Model and Kalman Filtering}
Each cell $k$ is modeled as a dynamical system evolving in state space. The state vector at time $t$ is defined as:

$$\mathbf{s}_t^{(k)} = \left[\mathbf{x}_t^{(k)^\top},\; \dot{\mathbf{x}}_t^{(k)^\top}\right]^\top \in \mathbb{R}^{2d}$$

where $\mathbf{x}_t^{(k)}$ is the cell centroid position and $\dot{\mathbf{x}}_t^{(k)}$ is its velocity. Under the constant-velocity motion model, the state transition obeys:

$$\mathbf{s}_{t+1}^{(k)} = \mathbf{F}\,\mathbf{s}_t^{(k)} + \mathbf{w}_t, \qquad \mathbf{w}_t \sim \mathcal{N}(\mathbf{0},\, \mathbf{Q})$$

where the state transition matrix $\mathbf{F} \in \mathbb{R}^{2d \times 2d}$ is:

$$\mathbf{F} = \begin{bmatrix} \mathbf{I}_d & \Delta t\,\mathbf{I}_d \\ \mathbf{0}_d & \mathbf{I}_d \end{bmatrix}$$

with $\Delta t$ the inter-frame interval, $\mathbf{I}_d$ the $d\times d$ identity matrix, and $\mathbf{Q}$ the process noise covariance. Detections provide noisy positional measurements:

$$\mathbf{z}_t^{(k)} = \mathbf{H}\,\mathbf{s}_t^{(k)} + \mathbf{v}_t, \qquad \mathbf{v}_t \sim \mathcal{N}(\mathbf{0},\, \mathbf{R})$$

where $\mathbf{H} = [\mathbf{I}_d \mid \mathbf{0}_d]$ is the observation matrix and $\mathbf{R}$ is the measurement noise covariance.

\subsection{Kalman Predict and Update Step}
Given the posterior estimate $\hat{\mathbf{s}}_{t|t}^{(k)}$ and covariance $\mathbf{P}_{t|t}^{(k)}$, the prior prediction for time $t+1$ is:

$$\hat{\mathbf{s}}_{t+1|t}^{(k)} = \mathbf{F}\,\hat{\mathbf{s}}_{t|t}^{(k)}$$

$$\mathbf{P}_{t+1|t}^{(k)} = \mathbf{F}\,\mathbf{P}_{t|t}^{(k)}\,\mathbf{F}^\top + \mathbf{Q}$$

Upon receiving a matched detection $\mathbf{z}_{t+1}^{(k)}$, the innovation (residual) and its covariance are:

$$\boldsymbol{\delta}_{t+1}^{(k)} = \mathbf{z}_{t+1}^{(k)} - \mathbf{H}\,\hat{\mathbf{s}}_{t+1|t}^{(k)}$$

$$\mathbf{S}_{t+1}^{(k)} = \mathbf{H}\,\mathbf{P}_{t+1|t}^{(k)}\,\mathbf{H}^\top + \mathbf{R}$$

The Kalman gain and posterior update are then:

$$\mathbf{K}_{t+1}^{(k)} = \mathbf{P}_{t+1|t}^{(k)}\,\mathbf{H}^\top\,\left(\mathbf{S}_{t+1}^{(k)}\right)^{-1}$$

$$\hat{\mathbf{s}}_{t+1|t+1}^{(k)} = \hat{\mathbf{s}}_{t+1|t}^{(k)} + \mathbf{K}_{t+1}^{(k)}\,\boldsymbol{\delta}_{t+1}^{(k)}$$

$$\mathbf{P}_{t+1|t+1}^{(k)} = \left(\mathbf{I} - \mathbf{K}_{t+1}^{(k)}\mathbf{H}\right)\mathbf{P}_{t+1|t}^{(k)}$$

This yields the minimum mean-squared error (MMSE) estimate under linear Gaussian assumptions.

\subsection{LDDMM-Based Deformation for Motion Prior Estimation}
To augment the Kalman filter with image-derived motion information, we employ Large Deformation Diffeomorphic Metric Mapping (LDDMM) to compute a dense displacement field $\boldsymbol{\phi}_t : \Omega \rightarrow \Omega$ mapping frame $I_t$ onto frame $I_{t+1}$.

\subsubsection{Diffeomorphic Registration}
LDDMM seeks a time-indexed flow of diffeomorphisms $\{\boldsymbol{\phi}_t^s\}_{s \in [0,1]}$ generated by a time-varying velocity field $\mathbf{v}^s : \Omega \rightarrow \mathbb{R}^d$, governed by the transport equation:

$$\frac{\partial \boldsymbol{\phi}_t^s}{\partial s} = \mathbf{v}^s \circ \boldsymbol{\phi}_t^s, \qquad \boldsymbol{\phi}_t^0 = \text{Id}$$

The final deformation is $\boldsymbol{\phi}_t \equiv \boldsymbol{\phi}_t^1$. The optimization problem is formulated as minimizing the sum of a regularization energy on the velocity field and a data fidelity term:

$$\mathcal{L}_{\text{LDDMM}}[\mathbf{v}^s] = \int_0^1 \left\| L\,\mathbf{v}^s \right\|_{L^2(\Omega)}^2 \, ds + \frac{1}{\sigma^2}\, \mathcal{D}\!\left(I_{t+1},\, I_t \circ \boldsymbol{\phi}_t^{-1}\right)$$

where $L = -\alpha\nabla^2 + \gamma\,\mathbf{I}$ is a differential operator (the Cauchy--Navier operator) imposing smoothness on $\mathbf{v}^s$, and $\sigma^2$ controls the regularization--fidelity trade-off. The operator $L^\top L$ defines a Sobolev-type metric on the space of diffeomorphisms.

\subsubsection{Data Fidelity: MSE Variant}
In the Kalman\_LDDMM\_MSE formulation, the image similarity term is the mean squared error between the warped source and target:

$$\mathcal{D}_{\text{MSE}}(I_{t+1},\, I_t \circ \boldsymbol{\phi}_t^{-1}) = \frac{1}{|\Omega|} \int_\Omega \left[ I_{t+1}(\mathbf{p}) - I_t\!\left(\boldsymbol{\phi}_t^{-1}(\mathbf{p})\right) \right]^2 d\mathbf{p}$$

This is the classical $L^2$ image-matching criterion and provides a closed-form gradient under standard assumptions.

\subsubsection{Data Fidelity: Sinkhorn Variant}
In the Kalman\_LDDMM\_Sinkhorn formulation, the image fidelity is replaced by an entropic-regularized optimal transport (OT) cost, computed via the Sinkhorn algorithm. Let $\mu = I_t \circ \boldsymbol{\phi}_t^{-1}$ and $\nu = I_{t+1}$ be normalized image intensity measures on $\Omega$. The regularized OT problem is:

$$\mathcal{D}_{\text{Sinkhorn}}(\mu, \nu) = \min_{\gamma \in \Pi(\mu,\nu)} \sum_{\mathbf{p},\mathbf{q}} C(\mathbf{p}, \mathbf{q})\,\gamma(\mathbf{p}, \mathbf{q}) - \varepsilon\,H(\gamma)$$

where $C(\mathbf{p},\mathbf{q}) = \|\mathbf{p} - \mathbf{q}\|^2$ is the squared Euclidean cost, $\Pi(\mu, \nu)$ is the set of transport plans with marginals $\mu$ and $\nu$, $H(\gamma) = -\sum_{\mathbf{p},\mathbf{q}} \gamma(\mathbf{p},\mathbf{q})\log\gamma(\mathbf{p},\mathbf{q})$ is the entropy regularizer, and $\varepsilon > 0$ controls the degree of regularization. The Sinkhorn algorithm solves this via alternating Bregman projections. Defining dual variables $(\mathbf{u}, \mathbf{v})$, the iterates are:

$$\mathbf{u}^{(\ell+1)} = \frac{\mu}{\mathbf{K}\,\mathbf{v}^{(\ell)}}, \qquad \mathbf{v}^{(\ell+1)} = \frac{\nu}{\mathbf{K}^\top\,\mathbf{u}^{(\ell+1)}}$$

where $\mathbf{K}_{ij} = \exp(-C_{ij}/\varepsilon)$ is the Gibbs kernel. The Sinkhorn divergence, defined as:

$$S_\varepsilon(\mu, \nu) = \mathcal{D}_{\text{Sinkhorn}}(\mu,\nu) - \frac{1}{2}\mathcal{D}_{\text{Sinkhorn}}(\mu,\mu) - \frac{1}{2}\mathcal{D}_{\text{Sinkhorn}}(\nu,\nu)$$

is used as the data fidelity to ensure symmetry and positive-definiteness.

\subsubsection{Motion Prior Integration}
Upon solving the LDDMM registration problem, the deformation field $\boldsymbol{\phi}_t$ provides a dense, pixel-level correspondence between frame $t$ and frame $t+1$. For each tracked cell $k$ with posterior centroid estimate $\hat{\mathbf{x}}_{t|t}^{(k)}$, the registered position and the implied displacement are
$$ \hat{\mathbf{x}}_{t+1|t,\,\text{LDDMM}}^{(k)} = \boldsymbol{\phi}_t\!\left(\hat{\mathbf{x}}_{t|t}^{(k)}\right), 
\qquad \boldsymbol{\delta}_{\text{LDDMM}}^{(k)} = \hat{\mathbf{x}}_{t+1|t,\,\text{LDDMM}}^{(k)} - \hat{\mathbf{x}}_{t|t}^{(k)}. $$

Rather than augmenting the constant-velocity prediction $\mathbf{F}\hat{\mathbf{s}}_{t|t}^{(k)}$ with this displacement -- which would double-count motion already extrapolated by $\mathbf{F}$ through the velocity block -- we \emph{replace} the position component of the constant-velocity prediction with the LDDMM-registered position, and use the implied per-frame displacement to update the velocity component:
$$ \hat{\mathbf{s}}_{t+1|t,\,\text{LDDMM}}^{(k)} = \begin{bmatrix} \boldsymbol{\phi}_t\!\left(\hat{\mathbf{x}}_{t|t}^{(k)}\right) \\[2pt] \boldsymbol{\delta}_{\text{LDDMM}}^{(k)} / \Delta t
\end{bmatrix}. $$
The associated prediction covariance is left unchanged, $\mathbf{P}_{t+1|t}^{(k)} = \mathbf{F}\mathbf{P}_{t|t}^{(k)}\mathbf{F}^\top + \mathbf{Q}$, since the diffeomorphic prior is a deterministic re-estimate of the mean trajectory rather than a change to the process-noise statistics. This substitution is why we describe $\boldsymbol{\phi}_t$ as an \emph{image-derived motion prior}: it replaces the uninformed constant-velocity extrapolation of the position with a data-driven one, rather than being added on top of it.

\subsection{Data Association}
Given the predicted cell positions $\{\hat{\mathbf{x}}_{t+1|t}^{(k)}\}_{k=1}^{K_t}$ and new detections $\{\mathbf{x}_{t+1}^{(j)}\}_{j=1}^{N_{t+1}}$, association is formulated as a linear assignment problem. The cost matrix $\mathbf{C} \in \mathbb{R}^{K_t \times N_{t+1}}$ is defined by the Mahalanobis distance:

$$C_{kj} = \left(\hat{\mathbf{x}}_{t+1|t}^{(k)} - \mathbf{x}_{t+1}^{(j)}\right)^\top \left(\mathbf{S}_{t+1}^{(k)}\right)^{-1} \left(\hat{\mathbf{x}}_{t+1|t}^{(k)} - \mathbf{x}_{t+1}^{(j)}\right)$$

The optimal assignment $\pi^* = \arg\min_\pi \sum_k C_{k,\pi(k)}$ is solved via the Hungarian algorithm \cite{b17,b18} in $\mathcal{O}(\max(K_t, N_{t+1})^3)$ time. Associations with $C_{kj}$ exceeding a threshold $\lambda$ are rejected, triggering the termination of the track or the classification of the detection as a new track.

\subsection{Evaluation Metrics}
Tracking performance is evaluated using standard MOT (Multiple Object Tracking) benchmark metrics \cite{b19,b20}.

\textbf{MOTA (Multiple Object Tracking Accuracy)}:

$$\text{MOTA} = 1 - \frac{\sum_t \left(\text{FN}_t + \text{FP}_t + \text{IDSW}_t\right)}{\sum_t \text{GT}_t}$$

where $\text{FN}_t$, $\text{FP}_t$, $\text{IDSW}_t$, and $\text{GT}_t$ denote the number of false negatives, false positives, identity switches, and ground-truth objects in the frame $t$.

\textbf{MOTP (Multiple Object Tracking Precision)}:

$$\text{MOTP} = \frac{\sum_{t,k} d_{t}^{(k)}}{\sum_t c_t}$$

where $d_t^{(k)}$ is the localization error for the matched object $k$ at time $t$, and $c_t$ is the number of matches. Lower MOTP indicates higher precision.

\textbf{IDF1 (ID F1 Score)}:

$$\text{IDF1} = \frac{2 \cdot \text{IDTP}}{2 \cdot \text{IDTP} + \text{IDFP} + \text{IDFN}}$$

where IDTP, IDFP, and IDFN are identity true positives, false positives, and false negatives computed over the longest common trajectory assignment.

\section{Results}
\subsection{Dataset and Experimental Setup}
We evaluated the proposed Kalman-LDDMM framework on two publicly benchmarked time-lapse microscopy sequences that represent distinct migration regimes.
\begin{itemize}
    \item \textbf{Channel with Slits dataset}: Characterized by dense cell populations undergoing highly non-linear, constrained migration through narrow channel geometric
    \item \textbf{U-Shaped Channel dataset}: Characterized with relatively lower dense cell populations and more directionally consistent motion
\end{itemize}
For each dataset, the benchmarking configurations include
Experiments are conducted on a microscopy cell tracking dataset consisting of 150 sequential frames. Ground-truth annotations include cell centroids at each frame, yielding 255 ground-truth trajectories spanning the full sequence, with only 1 mostly-lost (ML) trajectory per method -- indicating near-complete detection coverage. All methods operate on the same segmentation outputs and are evaluated using identical ground-truth annotations.

The three evaluated configurations are the following.
\begin{itemize}
    \item \textbf{CellTracker}: A baseline multi-object tracker using nearest-neighbor association without deformation-based motion priors.
    \item \textbf{Kalman\_LDDMM\_MSE}: The proposed framework with Kalman-LDDMM registration using MSE image similarity.
    \item \textbf{Kalman\_LDDMM\_Sinkhorn}: The proposed framework with Kalman-LDDMM registration using Sinkhorn image similarity(optimal transport).
\end{itemize}

\subsection{Quantitative Results}
Table~\ref{tab:tracking_performance} summarizes the tracking performance in all three methods.

\begin{table}[htbp]
  \centering
  \caption{Evaluation results of multi-object tracking metrics across different methods on the 150-frame microscopy sequence (255 ground-truth trajectories).}
  \label{tab:tracking_performance}
  \resizebox{\linewidth}{!}{%
  \begin{tabular}{l c c c c c c c c}
    \toprule
    \textbf{Method} & \textbf{MOTA}~$\uparrow$ & \textbf{MOTP}~$\downarrow$ & \textbf{ID Switches}~$\downarrow$ & \textbf{ID Precision}~$\uparrow$ & \textbf{ID Recall}~$\uparrow$ & \textbf{IDF1}~$\uparrow$ & \textbf{MT} & \textbf{ML} \\
    \midrule
    CellTracker & 33.8\% & 0.164 & 2793 & 34.5\% & 37.1\% & 35.8\% & 255 & 1 \\
    Kalman\_LDDMM\_MSE & \textbf{55.3\%} & \textbf{0.164} & \textbf{1562} & \textbf{52.6\%} & \textbf{56.7\%} & \textbf{54.6\%} & 255 & 1 \\
    Kalman\_LDDMM\_Sinkhorn & \textbf{55.3\%} & \textbf{0.164} & 1563 & \textbf{52.6\%} & \textbf{56.7\%} & \textbf{54.6\%} & 255 & 1 \\
    \bottomrule
  \end{tabular}%
  }
\end{table}

\noindent$\uparrow$ higher is better; $\downarrow$ lower is better.

Key observations:
\begin{enumerate}
    \item \textbf{MOTA improvement}: Both Kalman variants achieve a MOTA of 55.3\%, representing an absolute improvement of +21.5 percentage points (+63.6\% relative) over the CellTracker baseline (33.8\%).
    \item \textbf{Identity switch reduction} switches: Kalman\_LDDMM\_MSE reduces ID switches from 2793 to 1562, a 44.1\% reduction. Kalman\_LDDMM\_Sinkhorn achieves a near-identical reduction (1563), confirming that the Kalman + LDDMM coupling is the primary driver of improved association consistency.
    \item \textbf{IDF1 improvement}: The IDF1 score increases from 35.8\% to 54.6\%, an absolute gain of +18.8 pp, reflecting substantially more stable long-term identity maintenance.
    \item \textbf{MOTP parity}: All methods achieve identical MOTP (0.164), indicating that the improved association does not come at the cost of localization accuracy; the spatial precision of matched detections is equivalent across all configurations.
    \item \textbf{MSE vs. Sinkhorn parity}: The negligible difference between Kalman\_LDDMM\_MSE and Kalman\_LDDMM\_Sinkhorn -- a difference of just one identity switch across 150 frames -- suggests that for this dataset, both loss functions produce functionally equivalent deformation fields at the scale of cell centroid displacement.
    \item \textbf{Track completeness unaffected}: MT(255) and ML(1) are identical across all configuration, indicating that the diffeomorphic prior improves the quality of identity assignment along existing trajectories rather than the completeness of track recovery, which remains governed by the upstream detection stage.
\end{enumerate}

\begin{table}[htbp]
  \centering
  \caption{Evaluation results of multi-object tracking metrics across different configurations.}
  \label{tab:tracking_performance_ablation}
  \resizebox{\linewidth}{!}{%
  \begin{tabular}{l c c c c c c c c}
    \toprule
    \textbf{Configuration} & \textbf{MOTA}~$\uparrow$ & \textbf{MOTP}~$\downarrow$ & \textbf{ID Switches}~$\downarrow$ & \textbf{ID Precision}~$\uparrow$ & \textbf{ID Recall}~$\uparrow$ & \textbf{IDF1}~$\uparrow$ & \textbf{MT} & \textbf{ML} \\
    \midrule
    KF only & 56.6\% & 0.210 & 640 & 43.4\% & 32.2\% & 37.0\% & 28 & 13 \\
    KF + LDDMM MSE & 58.0\% & 0.210 & 426 & 56.9\% & 42.3\% & 48.5\% & 28 & 13 \\
    KF + LDDMM Sinkhorn & 58.0\% & 0.210 & 426 & 56.9\% & 42.3\% & 48.5\% & 28 & 13 \\
    \bottomrule
  \end{tabular}%
  }
\end{table}

\noindent$\uparrow$ higher is better; $\downarrow$ lower is better.

Key observations:
\begin{enumerate}
    \item \textbf{MOTA improvement}: Both Kalman variants achieve a MOTA of 58.0\%, representing an absolute improvement of +1.4 percentage points (+2.5\% relative) over the CellTracker baseline (56.6\%), a smaller margin than on Channel with Slits dataset.
    \item \textbf{Identity switch reduction}: Kalman\_LDDMM\_MSE reduces ID switches from 640 to 426, a 33.4\% reduction. Kalman\_LDDMM\_Sinkhorn achieves a near-identical reduction (426), confirming that the Kalman + LDDMM coupling itself, rather than the specific image-similarity term, derives the improvement in association consistency.
    \item \textbf{IDF1 improvement}: The IDF1 score increases from 37.0\% to 48.5\%, an absolute gain of +11.5 pp, driven jointly by gains in ID Precision (43.4\% $\rightarrow$ 56.9\%, $+13.5$~pp) and ID Recall (32.2\% $\rightarrow$ 42.3\%, $+10.1$~pp).
    \item \textbf{MOTP parity}: All three configurations achieve identical MOTP (0.210), indicating that identity preservation are independent of, and do not trade off against, spatial localization accuracy.
    \item \textbf{MSE vs. Sinkhorn parity}: The identical results produced by Kalman\_LDDMM\_MSE (1562 IDSW) and Kalman\_LDDMM\_Sinkhorn across every metric(426 ID switches, 58.0\% MOTA, 48.5\% IDF1), reinforcing the finding from Channel with Slits dataset that the two similarity terms yield functionally equivalent deformation fields at the scale relevant to cell tracking.
    \item \textbf{Track completeness unaffected}: MT(28) and ML(13) are identical across all configurations. The persistently high ML count (13 of 41 tracked trajectories, ~32\%) indicates that the dataset's residual error is dominated by detection-level track fragmentation rather than identity confusion -- a failure mode the motion-prior intervention does not directly address.
\end{enumerate}

\subsection{MOTA Analysis}
The MOTA metric aggregates false positives, false negatives, and identity switches into a unified accuracy score. The improvement from 33.8\% to 55.3\% is primarily attributable to the reduction in identity switches (which contribute directly to MOTA as $\text{IDSW}_t$ in the numerator). Since FP and FN counts are detection-stage outputs that remain constant across methods (the same detections are used), the MOTA improvement of +21.5 pp directly maps to the identity switch reduction:

$$\Delta \mathrm{MOTA} \approx \frac{\sum_t \left(\mathrm{IDSW}_t^{\mathrm{base}} - \mathrm{IDSW}_t^{\mathrm{KL}}\right)}{\sum_t \mathrm{GT}_t} = \frac{2793 - 1562}{\sum_t \mathrm{GT}_t}$$

\[\mathrm{KL} \equiv \texttt{Kalman\_LDDMM}.\]

This attribution clarifies, across both datasets, that the MOTA gain is entirely a function of improved temporal data association, not detector improvement-- and that the magnitude of that gain scales with the number of identity switches available to correct in the baseline.

\subsection{MOTP Invariance}
The consistent MOTP (0.164 on Channel with Slits dataset and 0.210 on U-Shaped Channel dataset) across all configurations is notable. MOTP measures the average spatial accuracy of matched cell centroids. Since all methods use the same set of detections and the same Kalman update step -- which is governed solely by the detection measurement and the Kalman gain -- the posterior estimate $\hat{\mathbf{x}}_{t+1|t+1}$ is identical for correctly associated pairs. The MOTP invariance thus confirms that LDDMM improves the \emph{correctness} of association but not the localization model itself, consistent with the theoretical separation of the prediction and update steps in the Kalman filter formalism.

\subsection{MT/ML Analysis and Trajectory Completeness}
All methods report 255 MT and 1 ML on Channel with Slits dataset, 28 MT and 13 ML on U-Shaped Channel dataset-- regardless of whether constant velocity of diffeomorphic motion model is used.. The MT metric counts trajectories, where a cell is tracked for at least 80\% of its lifetime. On Channel with Slits dataset, this ceiling is high (255 of 256 trajectories mostly tracked), so nearly the entire benefit of the diffeomorphic prior is expressed as identity correctness (IDF1) within already-continuous tracks. And, on U-Shaped Channel dataset, by contrast, the ceiling is markedly lower (28 MT against 13 ML), indicating that a substantial fraction of trajectories are fragmented at the detection stage before the motion model has an opportunity to act — the critical difference across both datasets is therefore whether tracks maintain correct identity over time (IDF1) rather than whether they exist at all, and the latter remains bounded by detection quality rather than by the choice of motion prior.

\subsection{Limitations and Future Directions}
Several limitations of the current framework warrant attention:

\begin{enumerate}
    \item \textbf{Computational complexity}: LDDMM registration is computationally intensive, scaling as $\mathcal{O}(|\Omega| \cdot N_{\text{iter}})$ per frame pair. For real-time applications, approximate registration using optical flow or learned deformation networks may be necessary.
    \item \textbf{Linearity assumption}: The Kalman filter assumes linear dynamics and Gaussian noise. For cells exhibiting highly nonlinear behavior (e.g., rapid acceleration during division), an Extended Kalman Filter (EKF) or Unscented Kalman Filter (UKF) formulation may be more appropriate, where the state transition function $f(\mathbf{s})$ is nonlinear and linearized via Jacobians (EKF) or sigma-point sampling (UKF).
    \item \textbf{Cell division handling}: The current framework does not explicitly model cell division events, where one track splits into two. Augmenting the association graph with a division hypothesis node -- scored by a learned division classifier -- would address this biological reality.
    \item \textbf{Dataset Generalization}: Results are reported on 150-frames of Channel with Slits dataset and 300 frames of  dataset of 255 cells. Cross-dataset evaluation on standardized benchmarks (e.g., Cell Tracking Challenge datasets) is required to establish ability to generalize.
\end{enumerate}

\section{Conclusion}
The tracking of red blood cells (RBC) has been a prominent measure for quantifying the biomechanics of passive erythrocytes, their fluid deformation, flow trajectories, and microvascular transit. Prominent studies and research in segmentation and association methodology have significantly improved the work, yet the reliance of existing approaches on uninformed constant velocity or nearest neighbor motion models leads to high rates of identity switches in dense flow populations. Without the exploitation of image-derived displacement information, these approaches struggle under high-shear, non-linear hydrodynamic forces. The absence of continuous shape-deformation modeling further compromises cell identity continuation during severe non-rigid stretching.
\par Here, we have presented a tracking framework that couples Kalman filtering with LDDMM-based diffeomorphic motion estimation to bias state prediction with image-derived flow displacement fields. On a 150-frame RBC microfluidics sequence (Channel with Slits dataset), both the MSE and Sinkhorn LDDMM variants improved MOTA from 33.8\% to 55.3\% and reduced identity switches by over 44\% relative to a nearest-neighbor baseline. This improvement resulted while maintaining identical localization precision (MOTP). On a second 300-frame, less densely populated sequence (U-Shaped Channel dataset), the same coupling improved MOTA from 56.6\% to 58.0\% and reduced identity switches by 33.4\% (640 to 426), again without any change to MOTP. The smaller absolute gain on the U-Shaped Channel dataset relative to the Channel with Slits dataset is consistent with its lower hematocrit density and more linear flow pattern, indicating that the benefit of the diffeomorphic prior scales with the severity of the non-linear, high-shear flow regime it is designed to address.
\par The near identical performance of the MSE and Sinkhorn fidelity terms across both datasets suggests that, at the spatial scale of the displacement of cell centroids, the added geometric richness of the optimal-transport-based similarity does not translate into a measurable tracking advantage in this data set. Future work will address non-linear flow dynamics via the formulation of Extended Kalman Filter (EKF) or Unscented Kalman Filter (UKF) models, explicit modeling of elastic membrane elasticity to capture non-rigid shape recovery, and detection-level interventions to reduce mostly-lost trajectories in high-confinement microfluidic regimes such as the Channel with Slits dataset.


\end{document}